\documentclass[10pt,twocolumn]{article}

\usepackage{cvpr}
\usepackage{times}
\usepackage{epsfig}
\usepackage{graphicx}
\usepackage{amsmath}
\usepackage{amssymb}
\usepackage{algorithm2e}
\usepackage{multirow}
\usepackage{tabularx}
\usepackage{textcomp}
\usepackage{eso-pic}
\usepackage{caption}
\usepackage{subcaption}

\usepackage[pagebackref=true,breaklinks=true,letterpaper=true,colorlinks,bookmarks=false]{hyperref}

\cvprfinalcopy 

\def\cvprPaperID{2} 

\ifcvprfinal\fi
\begin{document}

\title{MultiNet++: Multi-Stream Feature Aggregation and Geometric Loss Strategy for Multi-Task Learning }

\author{Sumanth Chennupati$^{1,3}$, Ganesh Sistu$^2$, Senthil Yogamani$^2$ and Samir A Rawashdeh$^3$\\
%
$^1$Valeo North America, $^2$Valeo Vision Systems, $^3$University of Michigan-Dearborn \\
{\tt\small\href{mailto:schenn@umich.edu}{schenn@umich.edu}}, {\tt\small \href{mailto:ganesh.sistu@valeo.com}{ganesh.sistu@valeo.com},   \href{mailto:senthil.yogamani@valeo.com}{senthil.yogamani@valeo.com},    \href{mailto:srawa@umich.edu}{srawa@umich.edu}}
}

\maketitle

\begin{abstract}
   Multi-task learning is commonly used in autonomous driving for solving various visual perception tasks. It offers significant benefits in terms of both performance and computational complexity. Current work on multi-task learning networks focus on processing a single input image and there is no known implementation of multi-task learning handling a sequence of images. In this work, we propose a multi-stream multi-task network to take advantage of using feature representations from preceding frames in a video sequence for joint learning of segmentation, depth, and motion. The weights of the current and previous encoder are shared so that features computed in the previous frame can be leveraged without additional computation. In addition, we propose to use the geometric mean of task losses as a better alternative to the weighted average of task losses. The proposed loss function facilitates better handling of the difference in convergence rates of different tasks. Experimental results on KITTI, Cityscapes and SYNTHIA datasets demonstrate that the proposed strategies outperform various existing multi-task learning solutions.       
\end{abstract}

\section{Introduction}

Multi-task learning (MTL) \cite{Caruana1997} aims to jointly solve multiple tasks by leveraging the underlying similarities between independent or interdependent tasks. It is perceived as an attempt to improve generalization by learning a common feature representation for multiple tasks. Improvements in prediction accuracy and reduced computation complexities are significant benefits of MTL. This allowed deployment of MTL in various applications in computer vision (especially scene understanding) \cite{8500504,8100062,Chennupati_2019}, natural language processing \cite{sanh2018hierarchical,Dong2015MultiTaskLF}, speech recognition \cite{7178814,7404849}, reinforcement learning \cite{Dewangan2018DiGradMR,7989250}, drug discovery \cite{ramsundar2015massively,liu2018exploration}, \etc. 

\begin{figure}[!t]
    \centering
    \includegraphics[width=\columnwidth]{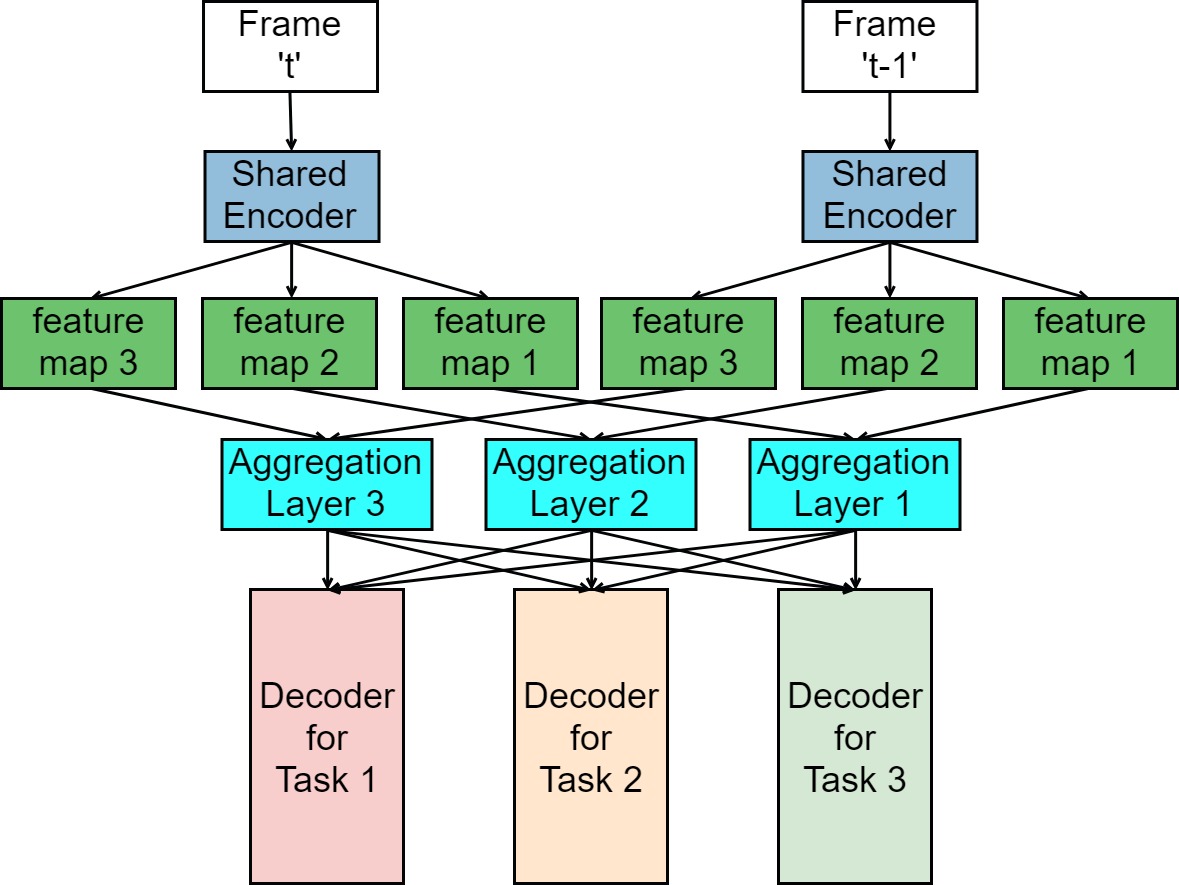}
    \caption{Illustration of MultiNet++ where feature aggregation is performed to combine intermediate output data obtained from a shared encoder that operates on multiple input streams (Frames `t' and `t-1'). The aggregated features are later processed by task specific decoders.  }
    \label{fig:multi-stream}
    \vspace{-0.4cm}
\end{figure}

MTL networks were mainly built using Convolution Neural Networks (CNNs). These networks were usually limited to operate on a single stream of input data. However, numerous works demonstrate using multiple streams of data as input to CNNs can improve performance drastically compared to using a single stream of input data. Recent attempts that use consecutive frames in a video sequence for semantic segmentation \cite{siam2017deep, Sistu_2019,Siam2017ConvolutionalGR}, activity recognition \cite{KarpathyCVPR14,simonyan2014two}, optical flow estimation \cite{ranjan2017optical}, moving object detection \cite{siam2018modnet, 8202211} are examples demonstrating the benefits of using multiple streams of input data. Similarly, a pair of images from stereo vision cameras \cite{ma2017multi} or multiple images from different cameras of a surround view system of a car can also be processed as multiple streams of input to CNNs. Some works considered processing input data from different domains \cite{sah} to solve certain tasks that require multi-modal data representations.

These significant benefits demand the construction of a multi-task learning network that can operate on multiple streams of input data. Thus, we propose MultiNet++, a novel multi-task network using simple feature aggregation methods as shown in Figure \ref{fig:multi-stream} to combine multiple streams of input data, which can be further processed by task-specific decoders. Figure \ref{fig:multi-stream} illustrates a generic way to aggregate features temporally and we make use of a simple summation junction to combine temporal features in our experiments. MultiNet++ would be ideal to process video sequences for tasks like semantic segmentation, depth estimation, optical flow estimation, object detection and tracking, \etc with improved efficiency. We also propose a novel loss strategy for multi-task learning based on geometric mean representation to prioritize learning of all tasks equally. 
The motivation for MultiNet++ is derived from our position paper NeurAll \cite{sistu2019neurall} which proposes to move towards a unified visual perception model for autonomous driving. We propose to use three diverse tasks namely segmentation, depth estimation and motion segmentation which make use of appearance, geometry and motion cues respectively.

The rest of the contents in this paper are structured as follows. Section \ref{sec:related-work} reviews related work using feature aggregation for multiple streams of inputs to CNNs and different task loss weighing strategies used in MTL. Section \ref{sec:multinet++} discusses in detail the proposed MultiNet++ network along with the geometric loss strategy used in this paper. Section \ref{sec:results} presents the experimental results on automotive datasets mainly KITTI \cite{Kitti}, Cityscapes \cite{Cordts2016Cityscapes} and SYNTHIA \cite{ros2016synthia}. Finally, Section \ref{sec:conclusion} summarizes the paper with key observations and concluding remarks.  

\section{Related Work}
\label{sec:related-work}
\subsection{Multi-Task Learning}
Multi-task learning typically consists of two blocks, shared parameters, and task-specific parameters. Shared parameters are learned to represent commonalities between several tasks while task-specific parameters are learned to perform independent processing. In MTL networks built using CNNs, shared parameters are called encoders as they perform the key feature extraction and the task-specific parameters are called decoders as they decode the information from encoders. MTL networks are classified into hard parameter sharing or soft parameter sharing categories based on how they share their parameters. In hard parameter sharing, initial layers or parameters are shared between different tasks such that these parameters are common for all tasks. In soft parameter sharing, different tasks are allowed to have different initial layers with some extent of sharing between them. Cross stitch \cite{Misra_2016} and sluice networks \cite{ruder2017learning} are examples of soft parameter sharing. Majority of the works in MTL use hard parameter sharing as it is easier to build and computationally less complex. 

The performance of the MTL network is highly dependent on their shared parameters as they contain the knowledge learned from different tasks \cite{Caruana1997,bilen2017universal,rebuffi2018efficient}. Inappropriate learning of these parameters can induce biased representations for a particular task which can hurt the performance of MTL networks. This phenomenon is referred to as negative transfer learning. In order to prevent it, meaningful feature representations and balanced learning methods are required.  

\subsection{Feature Aggregation}

Different outputs from initial or mid-level convolution layers from CNNs (referred to as extracted features) are forwarded to the next stage of processing using feature aggregation. Feature aggregation is a meaningful way to combine these extracted features. These features can be extracted from different CNNs operating on different input data \cite{Zhu_2017_ICCV,rashed2019optical}  or from a CNN operating on different resolutions of input \cite{lee2017multi}. Ranjan \etal \cite{Ranjan_2019} combines intermediate outputs from a CNN and passes to next stages of processing. Yu \etal \cite{Yu_2018_CVPR} proposed several possibilities of feature aggregation. 

There are plenty of choices to perform feature aggregation. These choices range from using simple concatenation techniques to complex Long Short Term Memory units (LSTMs) \cite{Hochreiter:1997:LSM:1246443.1246450} or recurrent units. Simple concatenation or addition layers can capture short term temporal cues from a video sequence.  Sun \etal \cite{Sun_2015} combine spatial and temporal features from video sequences for human activity recognition and Karpathy \etal \cite{KarpathyCVPR14} combine features from inputs separated by 15 frames in a video for classification. Hei Ng \etal \cite{Joe_Yue_Hei_Ng_2015} proposed several convolution and pooling operations to combine features for video classification while Sistu \etal \cite{Sistu_2019} used simple 1\texttimes1 bottleneck convolutions to combine features from consecutive frames for video segmentation. 

In automotive or indoor robotic visual perception problems, simple concatenation techniques perform well but they fall short in some applications like video captioning \cite{Donahue_2017,oruganti2016image} or summarization \cite{sah2017semantic} where long term dependencies are required. LSTMs in such cases offer a better alternative \cite{yao2015describing,sharmaaction}. Convolution-LSTMs (Conv-LSTMs) \cite{xingjian2015convolutional,Song_2018_ECCV} and 3D convolutions \cite{ji20133d} are other options. However, these options incur additional computational complexity and they are needed mainly for aggregation of features that are significant for long term dependencies.  

\begin{figure*}
    \centering
    \includegraphics[width=.9\textwidth]{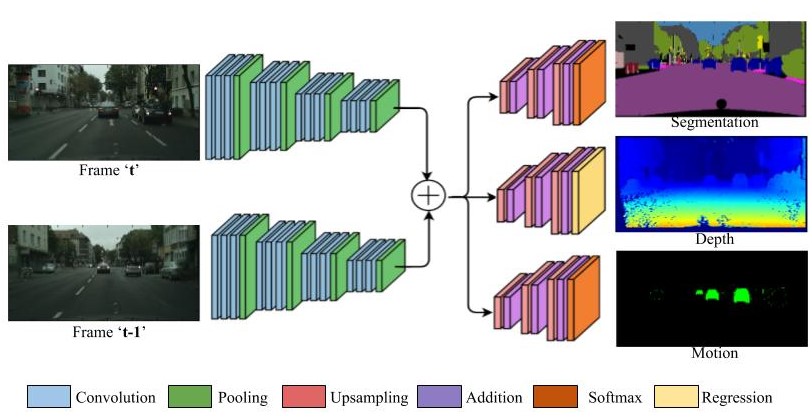}
    \caption{Illustration of the MultiNet++ network operating on consecutive frames of input video sequence. Consecutive frames are processed by a shared siamese-style encoder and extracted features are concatenated and processed by task specific segmentation, depth estimation and moving object detection decoders.  }
    \label{fig:multi-stream-task}
\end{figure*}

\subsection{Multi-Task Loss}
\label{sec:muti-tak-loss}
With the growing popularity of MTL, it is worth considering the possibility of imbalances in training an MTL network. It is often observed that some tasks dominate others during the training phase \cite{guo2018dynamic}. This dominance can be attributed to variations in task heuristics like complexities, uncertainties, and magnitudes of losses etc. Therefore an appropriate loss or prioritization strategy for all tasks in an MTL is a necessity. 

Early works in MTL \cite{8500504,8100062}, use a weighted arithmetic sum of individual task losses. Later, several works attempted to balance the task weights using certain task heuristics discussed earlier. Kendall \etal \cite{kendall2017multi} proposed to use homoscedastic uncertainty of tasks to weigh them. 
This approach requires explicit modeling of uncertainty and more importantly, the task weights remain constant.

GradNorm \cite{Chen2018GradNormGN} is another notable work in which Chen \etal proposes to normalize gradients from all tasks to a common scale during backpropagation. Lui \etal \cite{liu2018endtoend} proposed Dynamic Weight Average (DWA) which uses an average of task losses over time to weigh the task losses. Guo \etal \cite{guo2018dynamic} on the other hand proposed dynamic task prioritization where the changes in the difficulty of tasks adjust the task weights. This allows distributing focus on harder problems first and then on less challenging tasks. On another hand, Liu \etal devised a different strategy to use a reinforcement learning based approach to learn optimal task weights. However, this method isn't simple and it brings additional complexity to the training phase. 

In contrast to modeling multi-task problem as a single objective problem, Sener and Koltun \cite{sener2018multitask}  proposed to model it as a multi-optimization problem. Zhang and Yeung \cite{Zhang2010ACF} proposed a convex formulation for multi-task learning and Desideri \cite{desideri2012multiple} proposed a multiple-gradient descent algorithm. In summary, these strategies either involve an explicit definition of loss function using task heuristics or require complex optimization techniques. Therefore, a loss strategy with minimal design complexities will be well suited for multi-task learning to accommodate a virtually unlimited number of joint tasks.

\section{Proposed Solution}
\label{sec:multinet++}

We introduce our novel multi-task network MultiNet++, that is capable of processing multiple streams of input data. The proposed architecture is scalable and can be readily applied in any multi-task problem. In the following subsection, we discuss how we built our MultiNet++ network shown in Figure \ref{fig:multi-stream-task}.

\subsection{Multi-stream Multi-task Architecture}
\label{sec:arch-design}
MultiNet++ is a simple multi-task network with the ability to process multiple streams of input data. It is built using three main components, 1) Encoders that feed multiple streams of input into the network, 2) Feature aggregation layers that concatenate the encoded feature vectors from multiple streams and 3) Task-specific decoders that operate on aggregated feature space to perform task-specific operations. In this paper, we use MultiNet++ for joint semantic segmentation, depth estimation and moving object detection (or simply motion) on video sequences. We share the encoder between two consecutive frames from a given video sequence as shown in Figure \ref{fig:multi-stream-task}. This can significantly reduce the computational load as the encoders require a daunting number of parameters. These input frames can be selected sparsely or densely from a video sequence by observing its motion histogram. One can also choose to pass keyframes as proposed by Kulhare \etal \cite{7899739}. 

Our encoders are selected by removing fully connected layers from ResNet-50 \cite{7780459}. Outputs from ReLU \cite{hahnloser2001permitted} activation at layers 23, 39 and 46 from ResNet-50 \cite{7780459} encoder are extracted and sent to feature aggregation layers. These feature maps extracted from different streams of inputs are concatenated and sent to task-specific decoders as shown in Figure \ref{fig:multi-stream}.
Segmentation decoder is built using FCN8 \cite{long2015fully} architecture that comprises of 3 upsampling layers and skip connections from aggregated feature maps as shown in Figure \ref{fig:multi-stream-task}. The final layer consists of softmax \cite{Goodfellow-et-al-2016} units to predict pixel-wise classification labels. Similarly, we construct a motion decoder by changing the number of output classes in softmax units. Depth decoder is built by replacing softmax with regression units.  

\subsection{Geometric Loss Strategy}

We discussed the importance of a loss strategy that requires minimal effort during design phase in Section \ref{sec:muti-tak-loss}. The commonly used loss combination function is arithmetic mean and it suffers from differences in the scale of the individual losses. This is partially alleviated by weighted average of the losses but it is difficult to tune manually. We were motivated to explore geometric loss combination which is invariant to the scale of the individual losses. Thus we express the total loss of a multi-task learning problem as geometric mean of individual task losses. We refer to this as Geometric Loss Strategy (GLS). 
For an n-task problem with task losses $`\mathcal{L}_{1}$',`$\mathcal{L}_{2}$' $\dots$ `$\mathcal{L}_{n}$', we express total loss as: 
\begin{equation}
\label{eq:ALS}
 \displaystyle \mathcal{L}_{Total} = \prod_{i = 1}^n\sqrt[n]{\mathcal{L}_{i}}   
\end{equation}  For example, in a 3-task problem with losses `$\mathcal{L}_{1}$',`$\mathcal{L}_{2}$' and `$\mathcal{L}_{3}$', we express total loss:
\begin{equation}
\label{eq:ALS-3task}
    \mathcal{L}_{Total} =  \sqrt[3]{\mathcal{L}_{1}\mathcal{L}_{2}\mathcal{L}_{3}}
\end{equation} Equations \ref{eq:ALS} and \ref{eq:ALS-3task} are quite popular in geometric programming. 
This loss function is differentiable and can be optimized using an optimizer like Stochastic Gradient Descent (SGD). In fact, this definition makes sure that all tasks are making progress. 
We adapt our loss function to focus or give more attention to certain tasks by introducing Focused Loss Strategy (FLS) where we multiply geometric mean of losses of focused tasks to existing loss function. %
In this case, we define loss function with focus on $m$ $(m \le n) $ important tasks as: 
\begin{equation}
\label{eq:fALS}
     \displaystyle \mathcal{L}_{Total} = \prod_{i = 1}^n \sqrt[n]{\mathcal{L}_{i}}\times\prod_{j = 1}^ m\sqrt[m]{\mathcal{L}_{j}} 
\end{equation} 


Equation \ref{eq:fALS}
provides an opportunity to focus on important tasks in a multi-task learning problem. Here we assume that the tasks are ordered in terms of priority so that first $m$ tasks out of the total $n$ tasks gets higher weightage. 

Application of $log$ function converts the product of losses to sum of $log$ of individual losses and thus can be interpreted to be equivalent to normalizing individual losses and then adding them. However, it is computationally  complex to make use of $log$ function.

\section{Experiments and Results}
\label{sec:results}

In this section, we discuss the datasets used for evaluating the efficacy of the proposed models. Later, we discuss in detail how we constructed the proposed models and provide a complexity analysis of each. We also discuss the optimization strategies used during the training phase. Finally, we provide the results obtained along with a discussion.  

\subsection{Datasets}

KITTI \cite{Kitti}, Cityscapes \cite{Cordts2016Cityscapes} and SYNTHIA \cite{ros2016synthia} are popular automotive datasets. KITTI has annotations for several tasks including semantic segmentation, depth estimation, object detection, \etc. However, these annotations were done separately for each task and the input is not always common across the tasks. KITTI Stereo 2015 \cite{Menze2018JPRS,Menze2015ISA} dataset  provides stereo images for depth estimation. A subset of these images is labeled for KITTI semantic segmentation \cite{Kitti}. This dataset consists of 200 train images and 200 test images. Cityscapes \cite{Cordts2016Cityscapes} dataset provides both segmentation and depth estimation annotations for $\approx$ 3500 images. Motion labels for these datasets are provided by Vertens \etal \cite{8202211}. 
SYNTHIA \cite{ros2016synthia} is a synthetic dataset that provides segmentation and depth annotations for raw video sequences simulated in different weather, light conditions and road types. KITTI \cite{Kitti} and Cityscapes \cite{Cordts2016Cityscapes} provide segmentation labels for 20 categories while SYNTHIA \cite{ros2016synthia} dataset provides segmentation labels for 13 categories.
\begin{table}[ht]
\centering
\resizebox{\columnwidth}{!}{%
\begin{tabular}{l|ccc}\hline
Annotations & KITTI\cite{Kitti} & Cityscapes\cite{Cordts2016Cityscapes} & SYNTHIA\cite{ros2016synthia} \\\hline \hline
Segmentation & \checkmark & \checkmark & \checkmark \\
Depth & \checkmark & \checkmark & \checkmark \\
Motion & \checkmark & \checkmark & \texttimes  \\ \hline \hline
\# Train & 200 & 2,975 & 888 \\
\# Validation  & 200 & 500 & 787\\
\# Type  & Real & Real & Synthetic\\
\hline
\end{tabular}%
}
\caption{Summary of the automotive datasets used in our experiments.}
\label{tab:datests}
\end{table}

\begin{table*}[!t]
\normalfont
\centering
\resizebox{.95\textwidth}{!}{%
\begin{tabular}{l|ccccc|cccc}
\hline
\multirow{2}{*}{Method} & \multicolumn{5}{c|}{KITTI \& Cityscapes} & \multicolumn{4}{c}{SYNTHIA} \\ 
 & Encoder & Segmentation & Depth  & Motion  & Total & Encoder & Segmentation  & Depth & Total \\ \hline \hline
 \multicolumn{10}{c}{\textbf{1-Task Segmentation, Depth or Motion}}\\ \hline
1-Task & 23.58M & 0.18M & - & - & 23.77M & 23.58M & 0.14M & - & 23.68M \\
1-Task & 23.58M & - & 3.88K & - & 23.59M & 23.58M & - & 3.87K & 23.59M \\
1-Task & 23.58M & - & - & 8.33K & 23.60M & - & - & - & -  \\\hline \hline
\multicolumn{10}{c}{\textbf{2-Task Segmentation and Depth}} \\\hline
1-Frame & 23.58M & 0.18M & 3.88K & - & 23.77M & 23.58M & 95.34K & 3.88K & 23.69M \\
2-Frames & 23.58M & 0.26M & 7.46K & - & 23.86M &23.58M  &0.14M & 7.46K & 23.74M \\\hline \hline
\multicolumn{10}{c}{\textbf{2-Task Segmentation and Motion}} \\\hline
1-Frame & 23.58M & 0.18M & - & 8.33K & 23.78M & - & - & - & - \\
2-Frames  & 23.58M & 0.26M & - & 15.50K & 23.86M & - & - & - & - \\\hline \hline
\multicolumn{10}{c}{\textbf{3-Task Segmentation, Depth and Motion}} \\\hline
1-Frame & 23.58M & 0.18M & 3.88K & 8.33K & 23.79M & - & - & - & - \\
2-Frames & 23.58M & 0.26M & 7.46K & 15.50K & 23.87M & - & - & - & - \\\hline
\end{tabular}%
}
\caption{Comparative study: Parameters needed to construct 1-task segmentation, depth and motion, 2-task segmentation and depth, 2-task segmentation and motion and 3-task segmentation, depth and motion models. We compare 2-task and 3-task models that operate on 1-frame and 2-frames.} 
\label{tab:params}
\end{table*}

In KITTI \cite{Kitti} and Cityscapes \cite{Cordts2016Cityscapes} datasets, images are sampled and annotated sparsely from raw videos. This poses a challenge to approaches that use temporal methods for segmentation or motion detection tasks in videos. In addition to KITTI \cite{Kitti} and Cityscapes \cite{Cordts2016Cityscapes} datasets, we use SEQS-02 (New York-like city) and SEQS-05 (New York-like city) from SYNTHIA dataset for training and validation respectively in our experiments. These sequences provide segmentation and depth annotations for consecutive images in a video sequence. Thus they are more suitable for evaluating our multi-task model which operates on multiple streams of input data. Table \ref{tab:datests} provides a summary of different properties of the 3 datasets discussed so far.     
\begin{table*}[ht]
\normalfont
\centering
\resizebox{.95\textwidth}{!}{%
\begin{tabular}{l| ccc| ccc| cc}
\hline
\multicolumn{1}{c|}{\multirow{2}{*}{Method}} & \multicolumn{3}{c|}{KITTI} & \multicolumn{3}{c|}{Cityscapes} & \multicolumn{2}{c}{SYNTHIA} \\ 
 \multicolumn{1}{c|}{} & Segmentation & Depth & Motion & Segmentation & Depth & Motion & Segmentation & Depth \\\hline \hline
 \multicolumn{9}{c}{\textbf{1-Task Segmentation, Depth or Motion}} \\\hline 
1-Task & 81.74\% & - & - & 78.95\% & - & - & 84.08\% & - \\
1-Task & - & 75.91\% & - & - & 60.13\% & - & - & 73.19\% \\
1-Task & - & - &\textbf{98.49}\% & - & - & \textbf{98.72}\% & - & - \\  \hline \hline
\multicolumn{9}{c}{\textbf{2-Task Segmentation and Depth}} \\\hline
Equal weights & 74.30\% & 74.47\% & - & 73.76\% & 59.38\% & - &63.45\% & 71.84\% \\
GLS (ours) & 81.50\% & 74.92\% & - & 79.14\% & 60.15\% & - &86.87\% & 73.60\% \\
MultiNet++ & 81.01\% & 73.95\% & - & \textbf{83.07}\% & 60.15\% & - & \textbf{88.15}\% & \textbf{78.39}\% \\ \hline \hline
\multicolumn{9}{c}{\textbf{2-Task Segmentation and Motion}} \\\hline
Equal weights & 80.14\% & - & 97.88\% & 78.46\% & - & 98.25\% & - & - \\
GLS (ours) & 81.52\% & - & 97.93\% & 77.63\% & - & 98.83\% & - & - \\
MultiNet++ & 81.75\% & - & 98.15\% & 78.86\% & - & 98.65\% & - & - \\ \hline \hline
\multicolumn{9}{c}{\textbf{3-Task Segmentation, Depth and Motion}} \\\hline
Equal weights & 77.14\% & 76.15\% & 97.83\% & 72.71\% & 60.97\% & 98.20\% & - & - \\
GLS (ours) &\textbf{82.20}\% & \textbf{76.54}\% & 97.92\% & 77.38\% & 61.56\% & 98.72\% & - & - \\
MultiNet++& 80.06\% & 73.94\% & 97.94\% & 82.36\% & \textbf{62.74}\%  & 98.21\% & - & - \\\hline
\end{tabular}%
}
\caption{Improvements in learning segmentation, depth estimation and motion detection as multiple tasks using equal weights, proposed geometric loss strategy (GLS) and 2 stream feature aggregation with GLS (MultiNet++) vs independent networks (1-Task) on KITTI, Cityscapes and SYNTHIA datasets. }
\label{tab:main-results}
\end{table*}

\begin{figure*}[h]
    \centering
    \includegraphics[width=.9\textwidth,height=5.2cm,trim=0 0 0 5,clip]{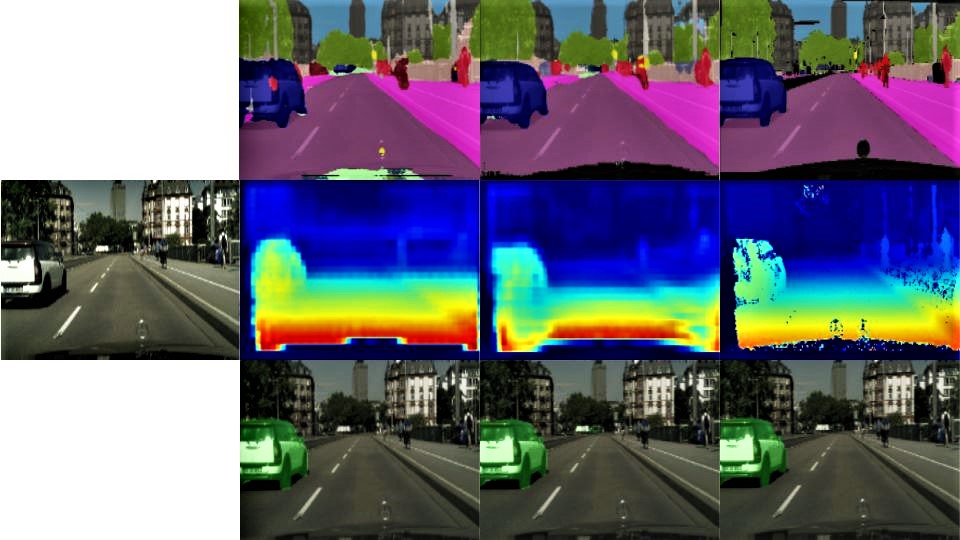}
    \includegraphics[width=.9\textwidth,height=5.2cm,trim=0 0 0 5,clip]{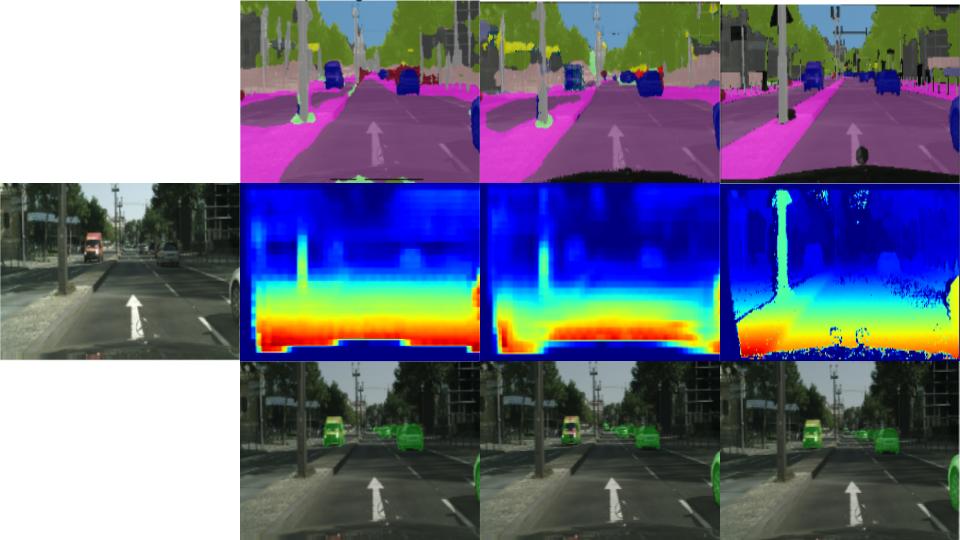}
    \centering
    \caption{Left to Right: Input Image, Single Task Network outputs, MultiNet++ Output, Ground Truth. More qualitative results of MultiNet++ model can be accessed via this link \url{https://youtu.be/E378PzLq7lQ}.}
    \label{fig:output }
\end{figure*}

\subsection{Model Analysis}

\begin{figure*}[h]
     \centering
     \begin{subfigure}[b]{0.3\textwidth}
         \centering \includegraphics[trim=25 25 5 5,clip,width=\textwidth]{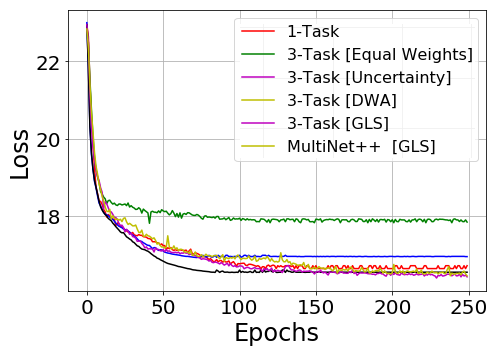}
         \caption{KITTI Segmentation}
         \label{fig:KITTI_Seg}
     \end{subfigure}
     \hfill
     \begin{subfigure}[b]{0.3\textwidth}
         \centering \includegraphics[trim=25 25 5 5,clip,width=\textwidth]{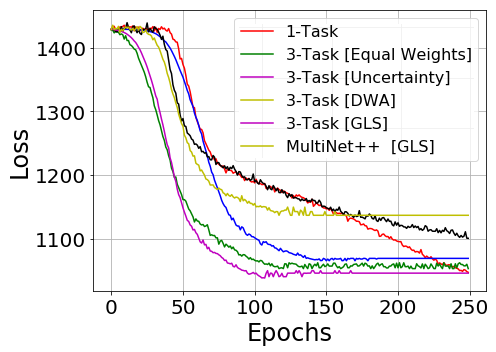}
         \caption{KITTI Depth}
        \label{fig:KITTI_Depth}
     \end{subfigure}
    \hfill
     \begin{subfigure}[b]{0.3\textwidth}
         \centering
         \includegraphics[trim=25 25 5 5,clip,width=\textwidth]{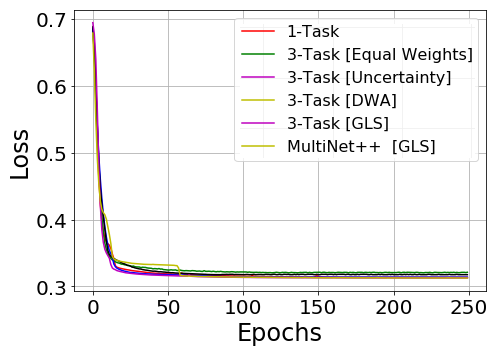}
         \caption{KITTI Motion}
         \label{fig:KITTI_Motion}
     \end{subfigure}
     \hfill
     \begin{subfigure}[b]{0.3\textwidth}
         \centering
         \includegraphics[trim=25 25 5 5,clip,width=\textwidth]{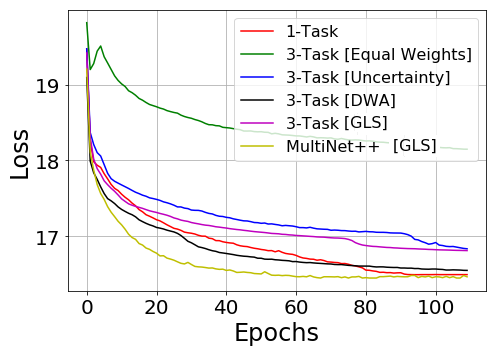}
         \caption{Cityscapes Segmentation}
         \label{fig:CITY_Seg}
     \end{subfigure}
     \hfill
     \begin{subfigure}[b]{0.3\textwidth}
         \centering
         \includegraphics[trim=25 25 5 5,clip,width=\textwidth]{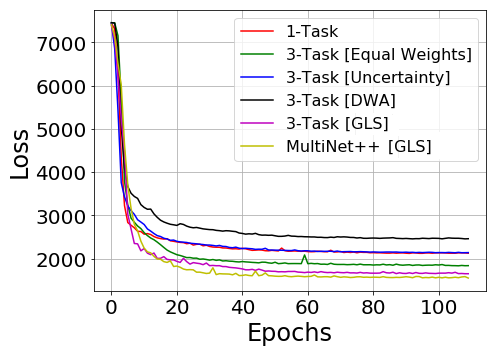}
         \caption{Cityscapes Depth}
         \label{fig:CITY_Depth}
     \end{subfigure}
     \hfill
     \begin{subfigure}[b]{0.3\textwidth}
         \centering \includegraphics[trim=25 25 5 5,clip,width=\textwidth]{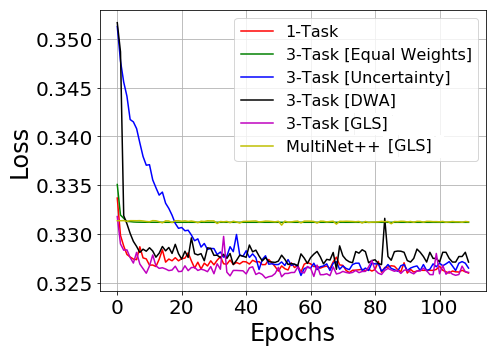}
         \caption{Cityscapes Motion}
 \label{fig:CITY_Motion}
     \end{subfigure}
        \caption{Change of validation loss (X-axis) over several epochs (Y-axis) during training phase for 1-Task model vs 3-Task models for segmentation, depth and motion tasks on KITTI \cite{Kitti} and Cityscapes \cite{Cordts2016Cityscapes} datasets.}
        \label{fig:Loss_graphs}
\end{figure*}

We constructed several models to evaluate the benefits of the proposed MultiNet++. We build 3 single task baseline models for segmentation, depth and motion tasks using ResNet-50 \cite{7780459} as an encoder and different task-specific decoders as discussed in Section \ref{sec:arch-design}. Segmentation decoder predicts pixel-wise labels from 20 different categories for input in KITTI \cite{Kitti} \& Cityscapes \cite{Cordts2016Cityscapes}  datasets, while the decoder predicts from 13 categories in SYNTHIA  \cite{ros2016synthia} dataset. Depth decoder outputs a 16-bit integer at every pixel location to predict depth and motion decoder predicts a binary classification label for every pixel to classify as moving or static object. These models process one frame of input data. We also constructed 2-task and 3-task models that operate on a single frame and 2 consecutive frames of an input video sequence. MultiNet++ refers to models that operate on 2 consecutive frames which are built using feature aggregation as discussed in Section \ref{sec:arch-design}. Table \ref{tab:params} provides details about number parameters required to construct different models.

Majority of computational load arises from ResNet-50 \cite{7780459} encoder. Due to this property, 2-task and 3-task models required the almost same number of parameters as 1-task model. This is one of the main reasons why multi-task networks are computationally efficient and favor embedded deployment. We build our 2-frame models with relatively very little increase in complexity ($\approx$  100K parameters) by reusing the encoder between 2-frames. In 2-frames model, the aggregated features are larger in size when compared to the 1-frame model. It resulted in an increase of parameters.

\subsection{Optimization}
We implemented our proposed models using Keras \cite{chollet2015keras}. In all our experiments, we re-size the input images to 224\texttimes384. We used only 2-frames for feature aggregation because adding more frames would increase computational complexity with insignificant performance gains as demonstrated by Sistu \etal \cite{Sistu_2019}.  
In our multi-task learning networks, we define the loss functions for each task separately and feed them to our geometric loss strategy (GLS) proposed in Section \ref{sec:muti-tak-loss}. 
For semantic segmentation and motion, we use pixel-wise cross-entropy loss for $C$ classes averaged over a mini-batch with $N$ samples as shown in Equation \ref{eq:seg-loss}.
\begin{equation}
    \label{eq:seg-loss}
    \displaystyle \mathcal{L}_{Seg} \text{ or } \mathcal{L}_{Motion} = -\sum_{j=1}^N\sum_{i=1}^C y_{i,j}log(p_{i,j})
\end{equation}

For depth estimation, we use Huber loss as defined in Equation \ref{eq:dep-loss} with $\delta$ =250.
\begin{equation}
    \label{eq:dep-loss}
     \mathcal{L}_{Depth} = \left\{
    \begin{array}{ll}
    \frac{1}{2} \left[y-\hat{y}\right]^2     & : |y-\hat{y}| \le \delta\\
    \delta \left(|y-\hat{y}|-\delta/2\right) & : otherwise
    \end{array}
    \right.
\end{equation}

The total loss $\mathcal{L}_{Total}$ is defined as:

\begin{equation}
    \label{eq:final-loss}
    \mathcal{L}_{Total} =  \sqrt[3]{\mathcal{L}_{Seg}\mathcal{L}_{Depth}\mathcal{L}_{Motion}}
\end{equation}

We optimize this loss function in our training phase using Adam optimizer \cite{kingma2014adam}. Accuracy is used as an evaluation metric for segmentation and motion tasks while regression accuracy is used for depth estimation. 
\begin{figure*}[h]
     \centering
     \begin{subfigure}[b]{0.16\textwidth}
         \centering \includegraphics[width=\textwidth,height=10cm]{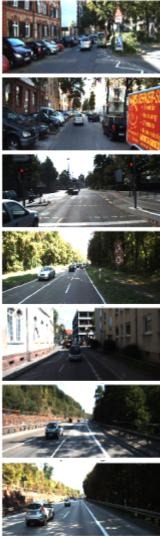}
         \caption{Input Images}
         \label{fig:KITTI_Input}
     \end{subfigure}
     \hfill
     \begin{subfigure}[b]{0.16\textwidth}
         \centering \includegraphics[width=\textwidth,height=10cm]{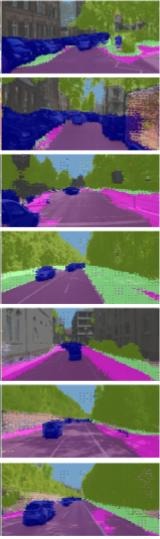}
         \caption{1-Task}
        \label{fig:Kitti_1task}
     \end{subfigure}
    \hfill
     \begin{subfigure}[b]{0.16\textwidth}
         \centering
         \includegraphics[width=\textwidth,height=10cm]{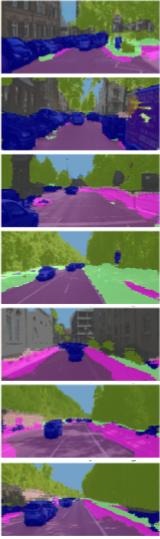}
         \caption{Equal Weights}
         \label{fig:3_taskequal}
     \end{subfigure}
     \hfill
     \begin{subfigure}[b]{0.16\textwidth}
         \centering
         \includegraphics[width=\textwidth,height=10cm]{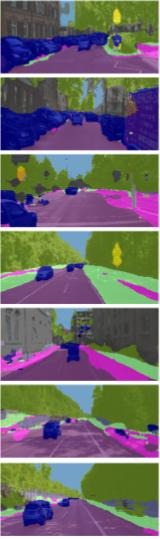}
         \caption{GLS}
         \label{fig:3_taskGLS}
     \end{subfigure}
     \hfill
     \begin{subfigure}[b]{0.16\textwidth}
         \centering
         \includegraphics[width=\textwidth,height=10cm]{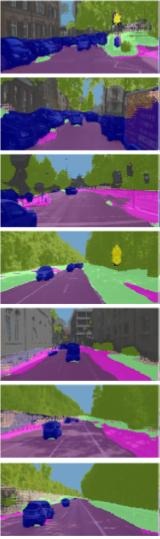}
         \caption{MultiNet++}
         \label{fig:3_taskmultinet++}
     \end{subfigure}
     \hfill
     \begin{subfigure}[b]{0.16\textwidth}
         \centering \includegraphics[width=\textwidth,height=10cm]{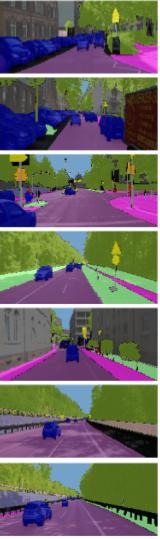}
         \caption{Ground Truth}
        \label{fig:Ground}
     \end{subfigure}
        \caption{Comparison of Semantic Segmentation results: 1-Task Segmentation vs 3-Task models on KITTI dataset.}
        \label{fig:Kitti_results }
\end{figure*}

\subsection{Results}

In Table \ref{tab:main-results}, we compare the results of 2-task models and 3-task models using our geometric loss strategy (GLS) against naive equal task weight method. We also compare their performances with 1-task segmentation, depth and motion models. Our GLS method shows significant improvements in performance over equal weights method in both 2-task and 3-task models.
In Table \ref{tab:comp-results}, we compare the results of 3-task models using our geometric loss strategy (GLS) against naive equal task weights, uncertainty weight method proposed by Kendal \etal \cite{kendall2017multi} and Dynamic Weight Average (DWA) proposed by Liu \etal \cite{liu2018endtoend}. In Figure \ref{fig:Loss_graphs} (\ref{fig:KITTI_Seg}, \ref{fig:KITTI_Depth}, \ref{fig:KITTI_Motion}, \ref{fig:CITY_Seg}, \ref{fig:CITY_Depth} and \ref{fig:CITY_Motion}), we show how validation loss for these models change over time during training phase. 
Our models using GLS demonstrated faster convergence on all tasks. 
\begin{table}[t]
\normalfont
\centering
\resizebox{.9\columnwidth}{!}{%
\begin{tabular}{l| ccc}
\hline
 \multicolumn{1}{c|}{Method} & Segmentation & Depth & Motion\\\hline \hline
\multicolumn{4}{c}{KITTI} \\\hline
1-Task & 81.74\% & 75.91\% & \textbf{98.49\%} \\
Equal weights & 77.14\% & 76.15\% & 97.83\%\\
Uncertainty \cite{kendall2017multi} &78.93\% & 75.73\% & 98.00\% \\
DWA \cite{liu2018endtoend} & 80.05\% & 74.48\% & 97.78\%\\
GLS (ours) &\textbf{82.20}\% & \textbf{76.54}\% & 97.92\%  \\
\hline \hline
\multicolumn{4}{c}{Cityscapes} \\\hline
1-Task & \textbf{78.95}\% & 60.13\% & \textbf{98.72}\% \\
Equal weights &  72.71\% & 60.97\% & 98.20\% \\
Uncertainty \cite{kendall2017multi} & 77.32\% & 60.44\% & 98.63\%\\
DWA \cite{liu2018endtoend} & 78.05\% & 59.34\% & 98.45\%\\
GLS (ours) & 77.38\% & \textbf{61.56}\% & \textbf{98.72}\%  \\
\hline
\end{tabular}%
}
\caption{Comparative Study: Performance of 1-Task, equal weights, 3-task uncertainty \cite{kendall2017multi}, Dynamic Weight Average (DWA) \cite{liu2018endtoend} and proposed geometric loss strategy (GLS) on KITTI and Cityscapes datasets. }
\label{tab:comp-results}
\end{table}
In 3-task models solving for segmentation, depth, and motion, depth is usually the most complex task. 
Figures \ref{fig:KITTI_Depth} and \ref{fig:CITY_Depth} show that depth estimation on KITTI \cite{Kitti} and Cityscapes \cite{Cordts2016Cityscapes} requires longer convergence time compared to segmentation (Figures \ref{fig:KITTI_Seg} and \ref{fig:CITY_Seg}) and motion tasks (Figures \ref{fig:KITTI_Motion} and \ref{fig:CITY_Motion}). In these cases, our GLS method has shown faster convergence compared to uncertainty \cite{kendall2017multi} and DWA \cite{liu2018endtoend} methods. While solving for multiple tasks, uncertainty \cite{kendall2017multi} and DWA \cite{liu2018endtoend} weigh the tasks that converge quickly higher than the others. This led to faster convergence in segmentation and motion tasks but late convergence in depth task. In such circumstances, the encoder parameters might be biased towards segmentation and motion tasks. This can result in imbalanced learning of depth task. Our GLS method expresses the total loss as the geometric mean of individual losses, so it doesn't prioritize one task higher than others. In this way, we achieve balanced training and improved performances compared to other techniques. 

In Table \ref{tab:main-results}, we also compare 2-task and 3-task models with our novel MultiNet++ which uses both feature aggregation (for 2-frame input) and GLS. In KITTI \cite{Kitti} dataset, input images are sparsely sampled from raw video sequences which hinder the performance gains of MultiNet++.  In Cityscapes \cite{Cordts2016Cityscapes} dataset, MultiNet++ outperforms single task models by 4\% and 3\% for segmentation and depth tasks respectively as they provide images sampled closely compared to KITTI dataset. These improvements are much better in SYNTHIA \cite{ros2016synthia} dataset (4\% and 5\% for segmentation and depth estimation tasks respectively) as they provide continuous frames of video sequences. We achieve similar performances for motion task compared to 1-task models. 

We compare qualitative results of MultiNet++ with 1-task segmentation model on Cityscapes \cite{Cordts2016Cityscapes} dataset in Figure \ref{fig:output }. The main difference between 1-task models and 3-task models is that the latter have learned representations from other tasks using a common encoder. Knowledge acquired through these representations helps 3-task model to identify semantic boundaries better compared to 1-task model. It is clearly evident that MultiNet++ model has improved performance. Our models detect traffic signs, lights and other near range objects better compared to other models on KITTI dataset \cite{Kitti} as shown in Figure \ref{fig:Kitti_results }.     

\section{Conclusion}
\label{sec:conclusion}
 
We introduced an efficient way of constructing MultiNet++, a multi-task learning network that operates on multiple streams of input data. We demonstrated that our geometric loss strategy (GLS) is robust to different task heuristics like complexity, magnitude, \etc. We achieved balanced training and improved performances for a multi-task learning network solving different tasks namely segmentation, depth estimation and motion on automotive datasets KITTI, Cityscapes, and SYNTHIA. Our GLS strategy is easy to implement and most importantly it allows for balanced learning of a large number of tasks in multi-task learning without requiring explicit loss modeling when compared to other multi-task learning loss strategies. In the future, we would like to explore the benefits of multi-task learning networks using our efficient feature aggregation and loss strategies for multi-modal data.

\section*{Acknowledgements}

Authors would like to thank their employer for supporting fundamental research. Authors would also like to thank Dr. Aditya Viswanathan and Dr. Thibault Julliand for helpful discussions.  
 

{\small
\bibliographystyle{ieee}
\bibliography{egbib}
}

\end{document}